\documentclass{llncs}
\usepackage{times,latexsym}
\usepackage{graphicx}
\usepackage{epsfig}
\usepackage{amsmath}
\usepackage{amssymb}

\begin{document}

\sloppy

\newcommand{\nin}{\noindent}

\newcommand{\ben}{\begin{enumerate}}
\newcommand{\een}{\end{enumerate}}
\newcommand{\ie}{\item}
\newcommand{\dist}{\operatorname{dist}}
\newcommand{\cost}{\operatorname{cost}}
\newcommand{\avg}{\operatorname{avg}}

\pagestyle{headings} 

\mainmatter

\title{Learning for Adaptive Real-time Search\vspace{-0.4cm}} 
 
\author{\vspace{-0.2cm}Vadim Bulitko}
\authorrunning{Vadim Bulitko}  
\tocauthor{Vadim Bulitko (University of Alberta)}
\institute{Department of Computing Science, University of Alberta, \\ 
Edmonton, Alberta T6G 2E8, Canada\\
\email{bulitko@ualberta.ca}\\ 
\texttt{http://www.cs.ualberta.ca/lrts}\vspace{-0.2cm}}

\maketitle 
 
\begin{abstract}\vspace{-0.6cm} 
Real-time heuristic search is a popular model of acting and learning in intelligent autonomous agents. Learning real-time search agents improve their performance over time by acquiring and refining a value function guiding the application of their actions. As computing the perfect value function is typically intractable, a heuristic approximation is acquired instead. 
Most studies of learning in real-time search (and reinforcement learning) assume that a simple value-function-greedy policy is used to select actions. This is in contrast to practice, where high-performance is usually attained by interleaving planning and acting via a lookahead search of a non-trivial depth.
In this paper, we take a step toward bridging this gap and propose a novel algorithm that (i) learns a heuristic function to be used specifically with a lookahead-based policy, (ii) selects the lookahead depth adaptively in each state, (iii) gives the user  control over the trade-off between exploration and exploitation.  
We extensively evaluate the algorithm in the sliding tile puzzle testbed comparing it to the classical LRTA* and the more recent weighted LRTA*, bounded LRTA*, and FALCONS. Improvements of 5 to 30 folds in convergence speed are observed.

\smallskip
\nin {\bf Keywords:} real-time heuristic search, planning and learning, on-line learning, adaptive lookahead search.\vspace{-0.2cm}
\end{abstract}

\vspace{-0.6cm}
\section{Problem Formulation} 
\vspace{-0.4cm}
Complete search methods such as A* and IDA*
\cite{IDAStar} produce optimal solutions with admissible heuristics. The price of optimality is the substantial running time often exponential in the problem dimension~\cite{Ratner86}. This limits the applicability of complete search
in tasks with large state spaces and limited time per action. 
The body of research on real-time search trades off optimality of solution for running time~\cite{Korf-AIJ}. Consequently, these techniques are widely applied to real-time path-planning, game-playing, control, and general decision-making. Most approaches to building boundedly rational~\cite{RussellWefald-book} real-time decision-making agents interleave lookahead-based deliberation and backing up the information to select an action~\cite{Korf-AIJ,RussellWefald-book,barto95}. Accordingly, most efforts to increase the rationality of such agents fall into three categories: (i) better hand-engineered and automatically derived heuristic functions~\cite{korf96finding}, (ii) various lookahead control, state pruning methods, and search extension techniques~\cite{RussellWefald-book}, and (iii) specialized hardware~\cite{DeepBlue}.

In this paper we will focus on the first two ways of increasing rationality of autonomous decision-making agents. Namely, we will consider the framework of {\em learning} in real-time heuristic search~\cite{Korf-AIJ,Ishida97} henceforth referred to as LRTS. It is an attractive model of decision-making in autonomous agents since it allows the agent to improve its performance over repeated trials in the same environment. Hence, another way to view LRTS is in the light of on-line reinforcement learning (RL)~\cite{barto95}. Not only the learning ability enables a gradual improvement of the solution, but also it allows the agent to act with an incomplete model of the environment, within non-stationary environment and goals, and in a non-deterministic environment. Consequently, the LRTS model has been successfully applied to numerous practical tasks including moving target search problems~\cite{MTS}, robot navigation and localization~\cite{Koenig98}, and robot exploration~\cite{Koenig99}.
Several important attributes of LRTS algorithms and RL agents are: 

\nin {\bf $\bullet$ final solution quality}  is measured in relation to the optimal solution quality. Having upper bounds on the cost of the solution the agent eventually converges to is important for performance guarantees;

\nin {\bf $\bullet$ convergence speed} is measured in the total number of actions the agent executes before it converges to its final solution; 

\nin {\bf $\bullet$ resource bounds} are imposed on the amount of memory the agent requires to converge to its final solution. This can be crucial for autonomous agents as their hardware is particularly limited;  

\nin {\bf $\bullet$ exploration versus exploitation control.} As finding high-quality solutions requires an extensive exploration of the state space, there is usually a trade-off between the final solution quality and the speed of the convergence and the resources required. Thus, an important attribute of an LRTS algorithm is whether this trade-off can be user-controlled;

\nin {\bf $\bullet$ convergence stability} stems from a related exploration and exploitation trade-off. Namely, fast convergence to better solutions requires an aggressive exploration of the state space. Such an LRTS agent usually demonstrates ``optimism in the face of uncertainty"~\cite{Moore93} by optimistically abandoning already found solution in an eager exploration of unknown regions of the state space. As a result, the solution quality can vary by orders of magnitude in consecutive trials. As argued in~\cite{ShimboIshida03}, this may be unacceptable depending on the application;

\nin {\bf $\bullet$ integration of learning and planning} is critical in most real-world LRTS (and learning game-playing) agents as their high performance crucially depends on extensive lookahead that constitutes the planning step of each cycle~\cite{DeepBlue,Chinook,ProbCut}. Contrary to a popular belief, deeper lookahead does not necessarily improve the decision-making quality~\cite{Nau-MiniMaxpathology,ijcai03-path,ecrts03-path}. Thus, an adaptive control of lookahead (i.e., planning)~\cite{RussellWefald-book} and its integration with the heuristic function update (i.e., learning) appear promising.

\vspace{-0.4cm}  
\section{Related Research}
\vspace{-0.4cm}

We will now consider several LRTS algorithms in light of the attributes discussed in the previous section. LRTA*~\cite{Korf-AIJ} is an early and still widely used LRTS algorithm. Under certain assumptions, it is guaranteed to convergence to an optimal solution in a finite number of trials. Being an optimality-seeking algorithm, LRTA* can require prohibitive computational resources (memory), is unstable in convergence~\cite{ShimboIshida03}, does not provide an exploration vs. exploitation control, and does not learn the heuristic function tailored to its lookahead module (called `mini-min' by Korf). 

Several extensions of LRTA* have been proposed. Weighted LRTA*~\cite{ShimboIshida03} is a combination of LRTA* with an inadmissible initial heuristic function. The heuristic is required to be within $(1+\epsilon)$ factor of the optimal heuristic function (i.e., $\epsilon$-admissible). The $\epsilon$ parameter controls the suboptimality of the final solutions as weighted LRTA* converges to a solution with the cost within $(1+\epsilon)$ of optimal. Consequently, larger values of $\epsilon$ lead to faster convergence and smaller memory requirements. The update rule is taken directly from LRTA* and does not take the lookahead into account. As in LRTA*, convergence stability is not directly addressed.

Bounded LRTA*~\cite{ShimboIshida03} uses additional memory to maintain an upper bound on the true heuristic function in each explored state. There is a user-set parameter $\delta$ used together with the upper bound to control the amount of exploration thereby increasing convergence stability.
On the other hand, bounded LRTA* incurs an additional memory and running time overhead as well as additional complexity of acquiring a non-trivial upper bound and refining it during the search. Additionally, its optimality seeking nature (for $\delta>2$) can lead to intractable storage requirements. Consequently, a combination of weighted and bounded LRTA* extensions is needed. The resulting interplay between $\epsilon$ and $\delta$ parameters and its effects on the algorithm properties is a subject of future research~\cite[p. 35]{ShimboIshida03}. Finally, there is still no explicit consideration of lookahead in the learning module of bounded LRTA* and it is not clear how well the weighting and upper-bounding techniques will work with the lookahead search of a non-trivial depth.

FALCONS~\cite{FALCONS} maintains an additional heuristic function which is a lower bound on the cost between the current state and the initial state. It was designed to speed up the convergence while retaining the optimality of the final solution and therefore can require prohibitive storage. Additionally, it does not offer the user any control over the exploration vs. exploitation trade-off or convergence stability. Finally, FALCONS does not make specific considerations of the lookahead in its learning module.

\vspace{-0.4cm}
\section{Novel Learning Real-time Search Algorithm} 
\vspace{-0.4cm}

\nin {\bf Intuition.} While past efforts have often focused {\em either} on gaining closer approximations to the true distance to goal~\cite{Korf-AIJ,korf96finding}
{\em or} better management of the planning phase~\cite{RussellWefald-book}, we recognize that these two are tightly coupled as the heuristic function is useful only insomuch as a guidance to the decision-making module of the agent. Consequently, instead of learning or manually engineering heuristic functions that approximate the true distance to goal well, we design the learning module specifically for the adaptive lookahead search. Below is the intuition behind the novel LRTS algorithm:
\ben\vspace{-0.2cm}
 
\ie Since optimal solutions are inherently intractable in many non-trivial domains (e.g.,~\cite{Ratner86}), we allow our algorithm to converge to suboptimal solutions. This results in significant speed-ups in convergence and savings in memory. At the same time, similar to weighted LRTA*, we upper-bound the converged solutions.

\ie Deep lookahead is wasteful in the states where the optimal action can be determined with a shallow search. Thus, our algorithm adjusts the lookahead depth adaptively based on the concept of so-called traps introduced later in the paper. Additionally, we execute a (short) variable-length sequence of actions per one lookahead thereby reducing the running time.

\ie We propose a tighter coupling between the heuristic function update and lookahead modules by updating the heuristic function only when the agent runs into a trap with respect to its lookahead module. This results in fewer heuristic function updates and smaller storage requirements.

\ie Similar to weighted LRTA*, we provide user with control over the optimality of the solutions and the resources required. The parameter can be selected based on the application at hand.

\een\vspace{-0.2cm}

\nin {\bf Traps.} A lookahead based real-time agent such as RTA* or LRTA* conducts lookahead of a fixed depth expanding a potentially large number of states at each step. Upon taking an action, it updates the heuristic function in its current state and repeats the process until the goal state is found.
Effectively, it follows the gradient of its heuristic function until it reaches a local minimum. 
This can be visualized as a `pit' in the heuristic function surface where the heuristic values of {\em all} surrounding states together with the distances to them exceed the heuristic value of the current state. An LRTA*-like agent will continue moving inside the `pit' and raising the heuristic values until the `pit' is completely `filled'. Such a `filling process' can take a large number of actions and updates to the heuristic function. We formalize such `pits' with a more general concept of $\gamma$-traps and incorporate explicit and efficient trap detection and recovery methods into our LRTS algorithm. 

\smallskip \nin {\bf Search problem.} Before we present the algorithm, the search task needs to be formally defined. An LRTS problem is defined as a tuple $(S,A,c,s_0,S_g)$ where $S$ is a finite set of states; $A$ is a finite set of actions; $c : S \times A \to (0,\infty)$ is the cost function with $c(s,a)$ being the incurred cost of taking action $a$ in state $s$; $s_0$ is the initial state; and $S_g \subset S$ is the set of goal states. We adopt the assumptions of~\cite{ShimboIshida03} that every action is reversible in every state, every applicable action leads to another state, and at least one goal state in $S_g$ is reachable from $s_0$. 

Then the true cost of traveling from state $s_1$ to state $s_2$ is denoted by $\dist(s_1,s_2)$ and is defined as the minimal cumulative action cost the agent is to incur by traveling from $s_1$ to $s_2$. For any state $s$ its true cost is defined as the minimal travel cost to the nearest goal: $h^*(s)=\min_{s_g \in S_g} \dist(s,s_g)$. A heuristic approximation $h$ to the true cost is called admissible iff $\forall s \in S \left[ h(s) \le h^*(s) \right]$. The value of $h$ in state $s$ will be referred to as the heuristic cost of state $s$.
Depth $d$ child of state $s$ is any state $s'$ reachable from $s$ in the minimum of $d$ actions (denoted by $\|s,s'\| = d$). Depth $d$ neighborhood of state $s$ is then defined as $S_{s,d} = \{s_d \in S \mid \|s,s_d\| = d\}$.
A state $s$ is called a $\gamma$-trap of depth $d$ iff it lacks a child of depth up to $d$ such that the cost of getting to that child adjusted by $\gamma$ plus the heuristic cost of the child does not exceed the heuristic cost of the current state: $\neg \left[ \exists d' \in \overline{1 \ldots d} \ \exists s' \in S_{s,d'} \left[ \gamma \cdot \dist(s,s') + h(s') \le h(s) \right]\right].$

\smallskip \nin {\bf The algorithm.} Figure~\ref{fig:gTrap} presents our LRTS algorithm called $\gamma$-Trap as a control policy outputting a series of  actions $\vec{a}$ in the state $s$. Thus, for a given search problem $(S,A,c,s_0,S_g)$ the agent's current $s$ state is initialized to $s_0$ and the heuristic function $h$ is initialized to an {\em admissible} initial function $h_0$. The $\gamma$-Trap policy is then called repeatedly by the environment and the series of actions $\vec{a}$ it returns are used to update the current state until a goal state is reached. During this process, the heuristic function $h$ is updated by the policy as well. Once a goal state is reached, the current problem-solving trial is deemed completed and the current state is reset to its initial location $s_0$. The agent is then to solve the problem again starting with the heuristic function $h$ from the previous problem-solving trial. The convergence is achieved when a problem-solving trial is completed without any updates to the heuristic function $h$. The solution produced (i.e., the sequence of actions the agent took from $s_0$ to the goal state) is considered to be the final solution. The cumulative cost of its actions is the final solution cost.  

\begin{figure}[htbp]
\vspace{-0.4cm}
\hrule\smallskip\textbf{$\gamma$-Trap}$(s,h,d_\text{max},\gamma)$

\vspace{1mm}

\textsc{Input}:

\vspace{1mm}

\begin{tabular} {l l}
\hspace{5mm} $s$: & the current state \\
\hspace{5mm} $h$: & heuristic function \\
\hspace{5mm} $d_\text{max}$: & maximum lookahead depth \\
\hspace{5mm} $\gamma$: & solution quality control parameter \\
\end{tabular}

\vspace{2mm}

\textsc{Output}: a series of actions $\vec{a}$

\vspace{2mm}

\begin{tabular} {p{0.5cm}l}
1 &  {\bf for} $d=1$ to $d_\text{max}$ do \\
2 &  \hspace{5mm} generate depth $d$ neighborhood $S_{s,d} = \{s_d \in S \mid \|s,s_d\| = d\}$ \\
3 &  \hspace{5mm} {\bf if} $s$ is {\em not} a $\gamma$-trap of depth $d$: i.e., $\exists s' \in S_{s,d} \left[  \gamma \cdot \dist(s,s') + h(s') \le h(s) \right]$ \\
4 & \hspace{10mm} {\bf return} actions $\vec{a}$ leading from $s$ to $\arg \min\limits_{s_d \in S_{s,d}} \left( \gamma \cdot \dist(s,s_d) + h(s_d)\right)$\\
5 &  {\bf end for loop} \\
6 &  update the heuristic function: $h(s) \leftarrow \max\limits_{d =1,\ldots,d_\text{max}} \; \min\limits_{s_d \in S_{s,d}} \left( \gamma \cdot \dist(s,s_d) + h(s_d)\right)$ \\
7 & {\bf return} actions $\vec{a}$ leading from $s$ to the previous current state \\
\end{tabular}

\smallskip\hrule
\caption{LRTS algorithm $\gamma$-Trap expressed as agent control policy.}\label{fig:gTrap}
\vspace{-0.6cm} 
\end{figure} 

$\gamma$-Trap operates as follows. Lines 1 through 5 conduct the lookahead search by incrementally expanding the neighborhood of the current state $s$ up to the maximum depth $d_\text{max}$ specified by the user. At each iteration of the expansion process if the current state $s$ is found to be a {\em non}-trap (line 3) then the lookahead search is terminated and the {\em series} of actions $\vec{a}$ leading to the most promising state within the last expanded neighborhood is returned (line 4). Ties are resolved randomly. Note that this operation exits the policy call. The agent will then move to the new current state $s$. The heuristic function $h$ will {\em not} be updated. On the other hand, if the loop terminates naturally in line 6 then the current state $s$ is a $\gamma$-trap of depth $d_\text{max}$. In this case the heuristic function needs to be updated. In order to minimize (i) the number of updates and (ii) the number of actions the agent takes while `filling the pit', we do two things. First, we increase the heuristic $h(s)$ by the maximum allowable amount in line 6 (hence $\max$ over $d$-neigborhoods). Second, we backtrack to the previous state (line 7) and avoid wasteful wandering inside the `pit'. Note that if the current state $s$ is the initial state $s_0$ no backtracking is performed and the current state is not changed. 

\smallskip\nin{\bf Completeness \& final solution quality.} Despite giving up final solution optimality, $\gamma$-Trap is complete and converges to a reasonable final solution. More precisely, the amount of optimality loss can be upper-bounded:

\smallskip\nin{\bf Theorem.} {\em
For any search problem $(S,A,c,s_0,S_g)$ that satisfies the aforementioned conditions and given an admissible initial heuristic $h_0$, $\gamma$-Trap will find a finite solution on every trial. Additionally, over a finite number of repeated trials, $\gamma$-Trap will converge to a final solution of the cost upper-bounded by $\frac{h^*(s_0)}{\gamma \min\limits_{S \times A} c(s,a)}$. 
}

\nin The proof is available on our website and cannot be reproduced here due to the space constraints. Also note that the actual performance of $\gamma$-Trap is substantially better than the upper bound suggests. For instance, in the convergence experiments detailed below the average final solution cost for $\gamma = 0.2$ was only $144\%$ of the optimal which is substantially below the upper bound of $500\%$ suggested by the theorem.

\vspace{-0.4cm}
\section{Empirical Evaluation}
\vspace{-0.4cm}

We evaluated the performance of $\gamma$-Trap against existing algorithms: weighted LRTA* due to Shimbo and Ishida~\cite{ShimboIshida03} and LRTA* and RTA* due to Korf~\cite{Korf-AIJ} which we had re-implemented. Additionally, since we extended the experimental setup used in~\cite{ShimboIshida03}, the figures for FALCONS and bounded LRTA* reported there are used in the comparisons as well.\footnote{NB: some of the figures reported in~\cite{ShimboIshida03} are incorrect due to a bug in their code. We thus use the figures recently re-computed by Shimbo~\cite{ShimboPC}.}
Our extensions to the experiments reported in~\cite{ShimboIshida03} were as follows: (i) we used up to 10 folds in order to measure the standard deviation and (ii) the algorithms were run with the lookahead depth fixed at 2, 5, 10, and 15 plies in addition to the lookahead of one reported therein. Due to the extensive nature of the results and the space constraints of the paper, only a subset of the findings will be reported herein. Full results can be found in a technical report on our web site.

LRTA* and RTA* were reimplemented directly from~\cite{Korf-AIJ}. Additionally, a version of LRTA* upgraded with full hash-based lookahead pruning, consistent use of table-based $h$ function, forced monotonicity of $h+g$ values, and only upward updates to $h$ (as in~\cite{ShimboIshida03}) was implemented and run under the name of iLRTA*. It was used for weighted LRTA* experiments and is listed as eiLRTA-x.x in the figures below where x.x is the value of $\epsilon$. Likewise, bounded LRTA* is listed as biLRTA-x.x in the graphs (x.x being the value of $\delta$).
In order to investigate the impact of backtracking in $\gamma$-Trap, we also implemented a version of the algorithm with no backtracking upon $h$-function update. Instead, the algorithm always moves into the child node with the minimum $h+\gamma \dist$ value. In the plots, the backtracking version of $\gamma$-Trap is listed as gTrap-BT-x.x whereas the non-backtracking version is listed as gTrap-x.x (x.x is the value of $\gamma$).
Manhattan distance was used as the initial heuristic function $h_0$ for all algorithms except weighted LRTA* where we used Manhattan distance multiplied by $(1+\epsilon)$.

\begin{figure}
\caption{Performance over repeated trials with the lookahead of one. Each number is averaged over 10 independent folds of 100 random 8-puzzle instances each. Each 8-puzzle was solved until convergence. The total costs are plotted in the top graph while the final solution costs relative to optimal are below. The data for bounded LRTA* and FALCONS are taken from~\cite{ShimboIshida03} and computed over only one fold (hence no error bars or final solution costs). }\label{fig:8conv-1}
\begin{center}
\includegraphics[width=12.5cm]{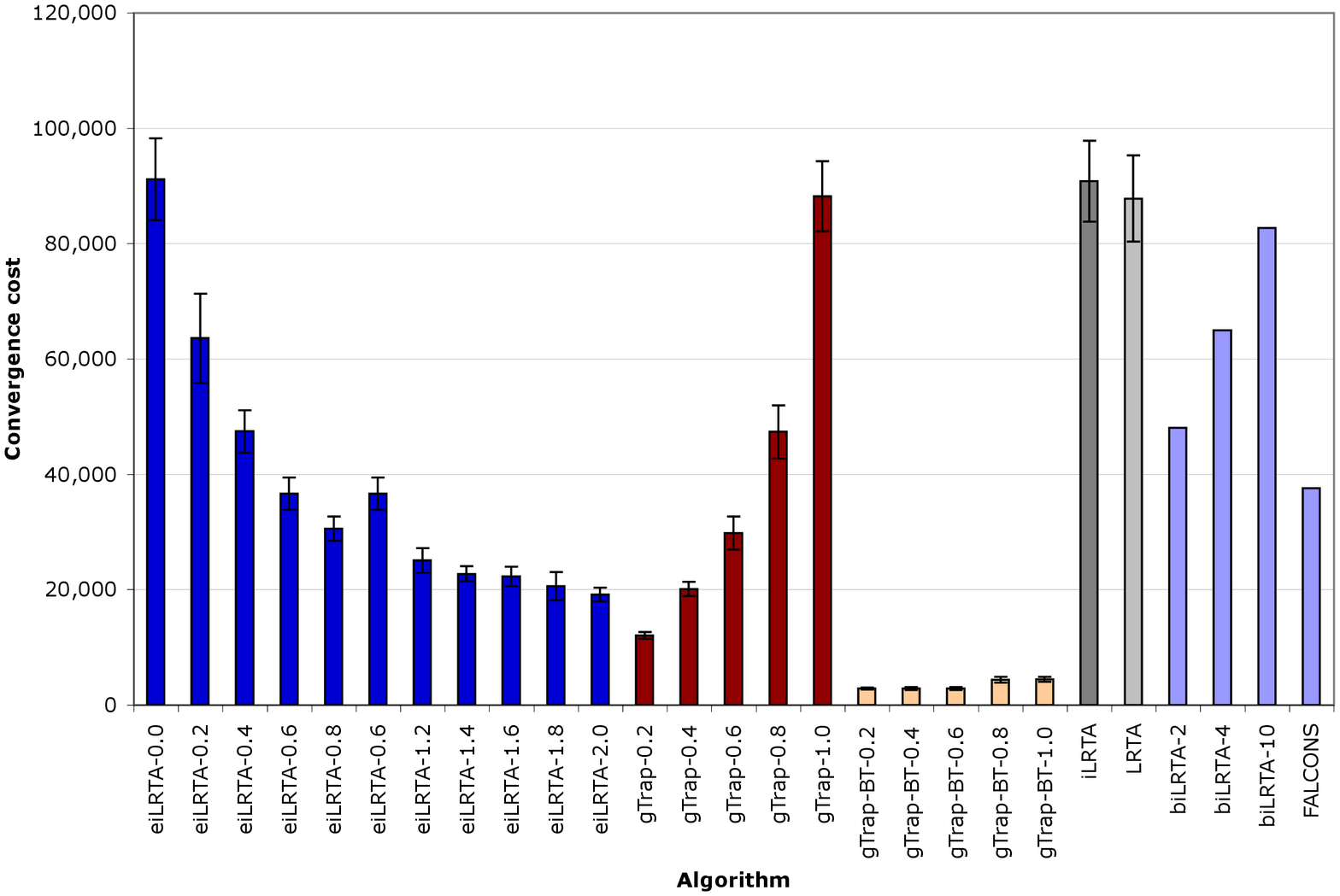}

\includegraphics[width=12.5cm]{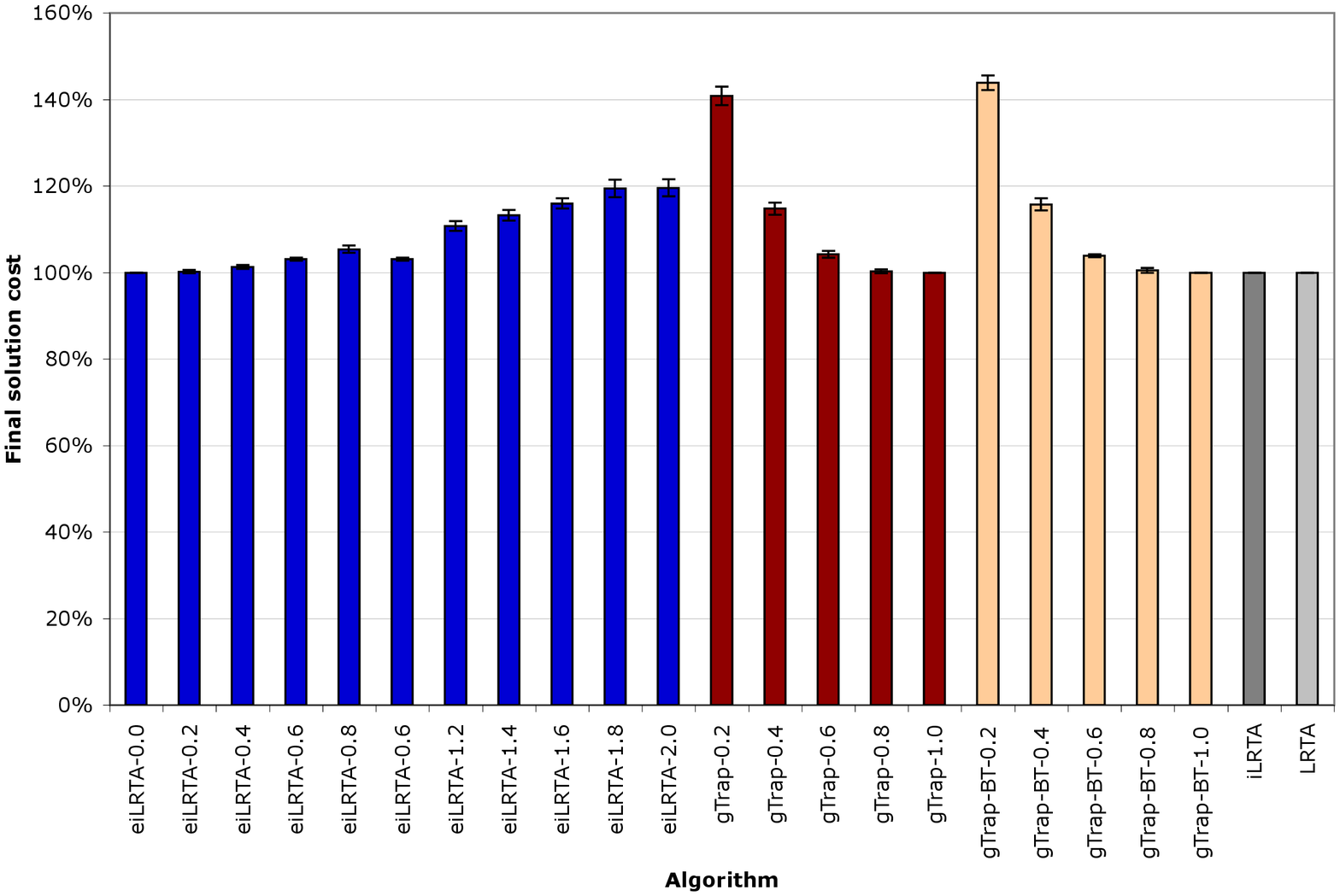}
\end{center}
\end{figure}

\medskip\nin{\bf Convergence Speed.} Within each fold, a random set of one hundred 8-puzzles was generated. Each of the algorithms (with the exception of non-learning RTA*) was run repeatedly on each of the puzzle instances until a convergence was achieved (indicated by the lack of $h$-function updates between the start and the goal states). The convergence cost and the final solution cost relative to the optimal were averaged over 100 puzzles. The the mean and standard deviation were computed over 10 folds of 100 puzzles each. The experiments were repeated for the lookahead depths of 1, 2, 5, 10, and 15.  The results for the lookahead of one are shown in Figure~\ref{fig:8conv-1}. As the data indicate, $\gamma$-Trap with backtracking (gTrap-BT) outperforms all other algorithms in terms of the convergence cost: approximately seven times better than the most successful weighted LRTA*, 13 times better than FALCONS, and 31 times better than classic LRTA*. At the same time, it is comparable to others in terms of the final solution cost: 144\% of optimal vs. 120\% in the case of the most successful weighted LRTA*. This trend was observed for all lookahead depths.




\begin{figure}
\caption{Top: the number of Korf's 15-puzzles solved to convergence within four million stored heuristic values. Bottom: the final solution quality attained. The lookahead is fixed at one.}\label{fig:15nc-1}
\begin{center}
\includegraphics[width=12.5cm]{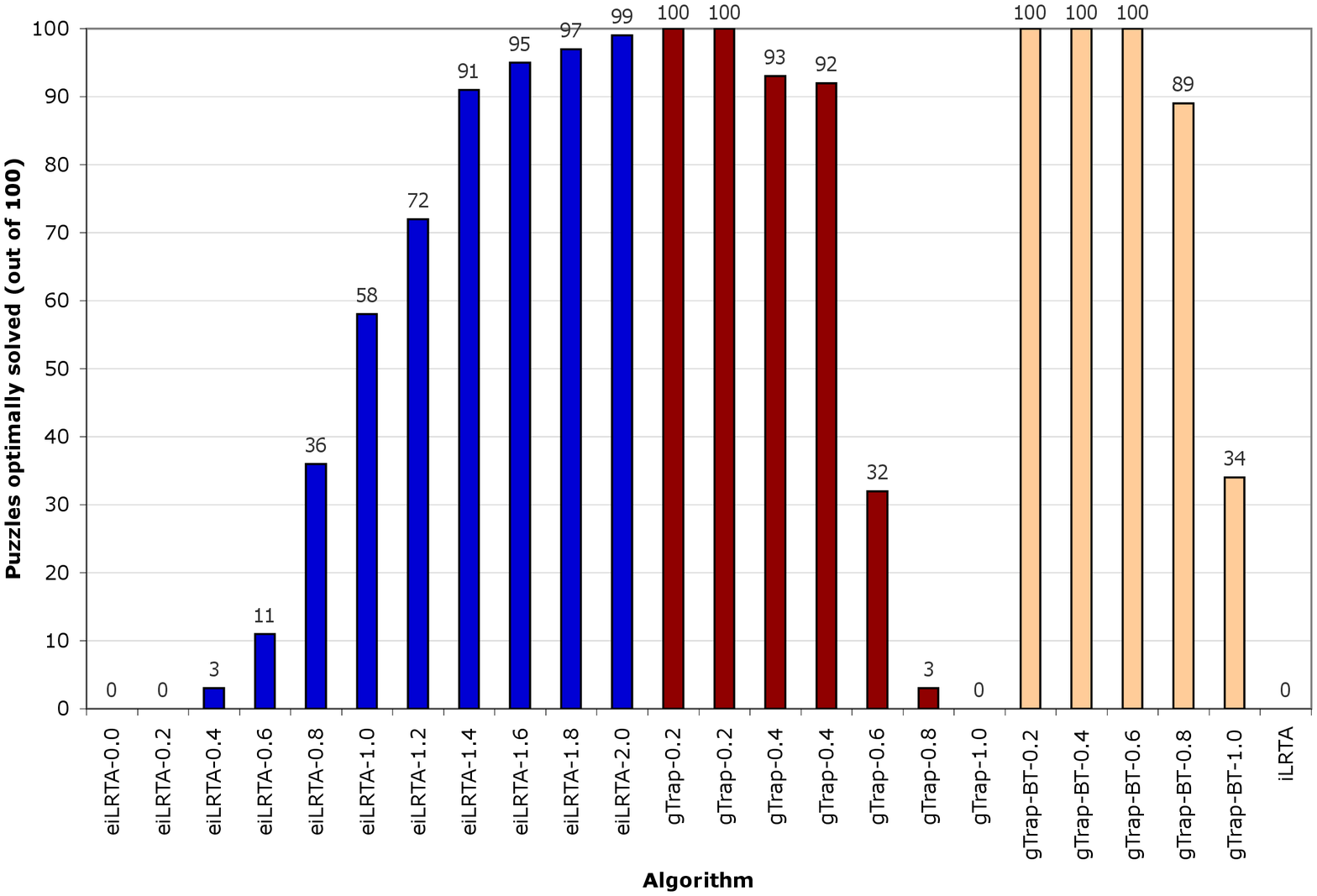}

\includegraphics[width=12.5cm]{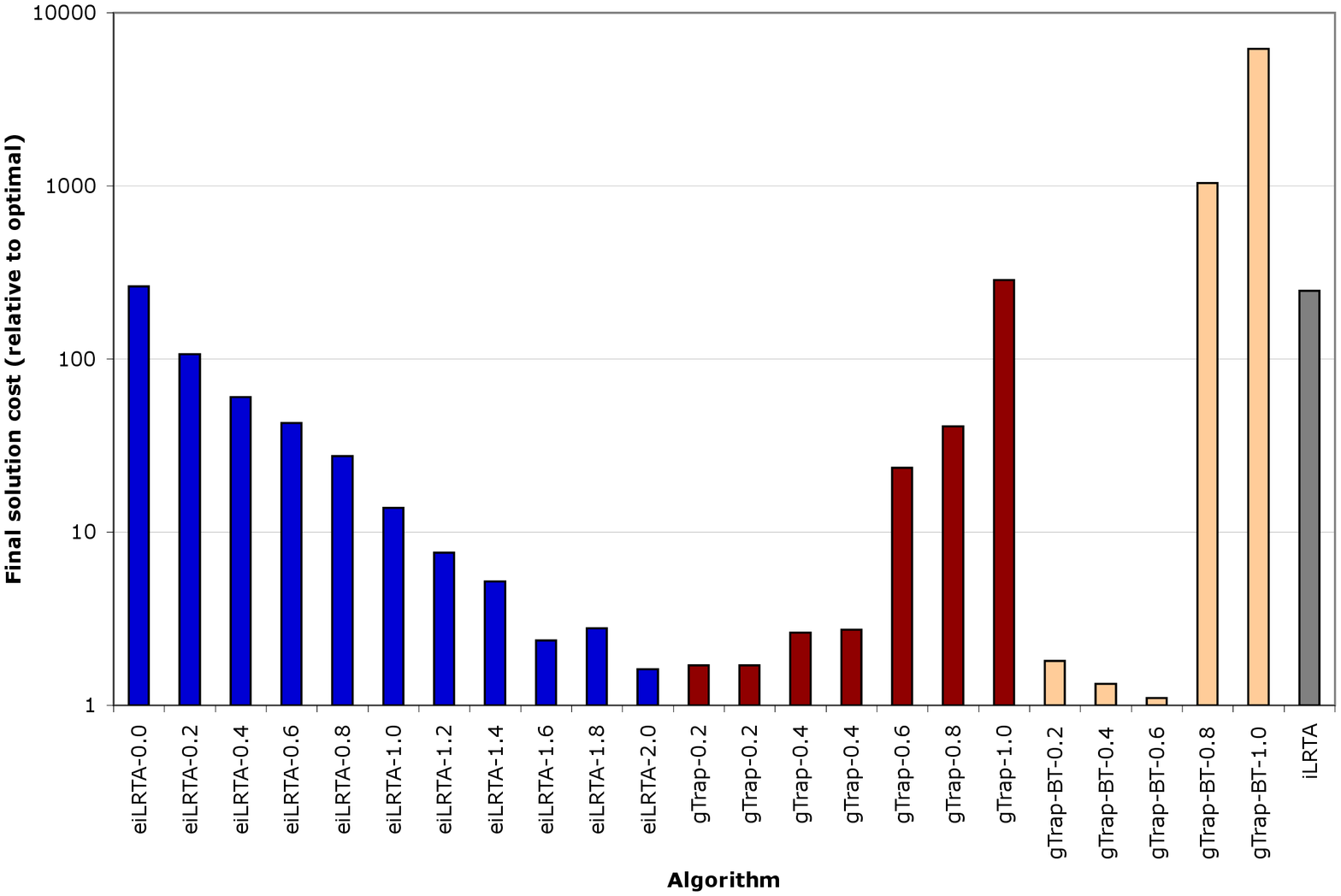}
\end{center}
\end{figure}

\medskip\nin{\bf Memory Requirements.} We paralleled the experiments reported in~\cite{ShimboIshida03} and computed the number of 15-puzzles from the one hundred puzzle set published by Korf in~\cite{IDAStar} that can be solved until convergence within four million heuristic function values stored. The results can be found in Figure~\ref{fig:15nc-1} for the lookahead of one. In the experiments, $\gamma$-Trap was the only algorithm converging on all 100 Korf's puzzles within four million nodes. At the same time, it has found the highest quality solutions among the algorithms (e.g., 110\% of optimal vs. 520\% of optimal with the most successful weighted LRTA*). 

\begin{figure}
\caption{Learning stability over 10 folds of 100 hundred random 8-puzzles each with the lookahead of one. SOD (top) and IAE (bottom) stability indices 
are plotted. The data for bounded LRTA* and FALCONs are from~\cite{ShimboIshida03}, were collected over a single fold, and thus do not have standard deviations plotted as the error bars.}\label{fig:8stab-1}
\begin{center}
\includegraphics[width=12.5cm]{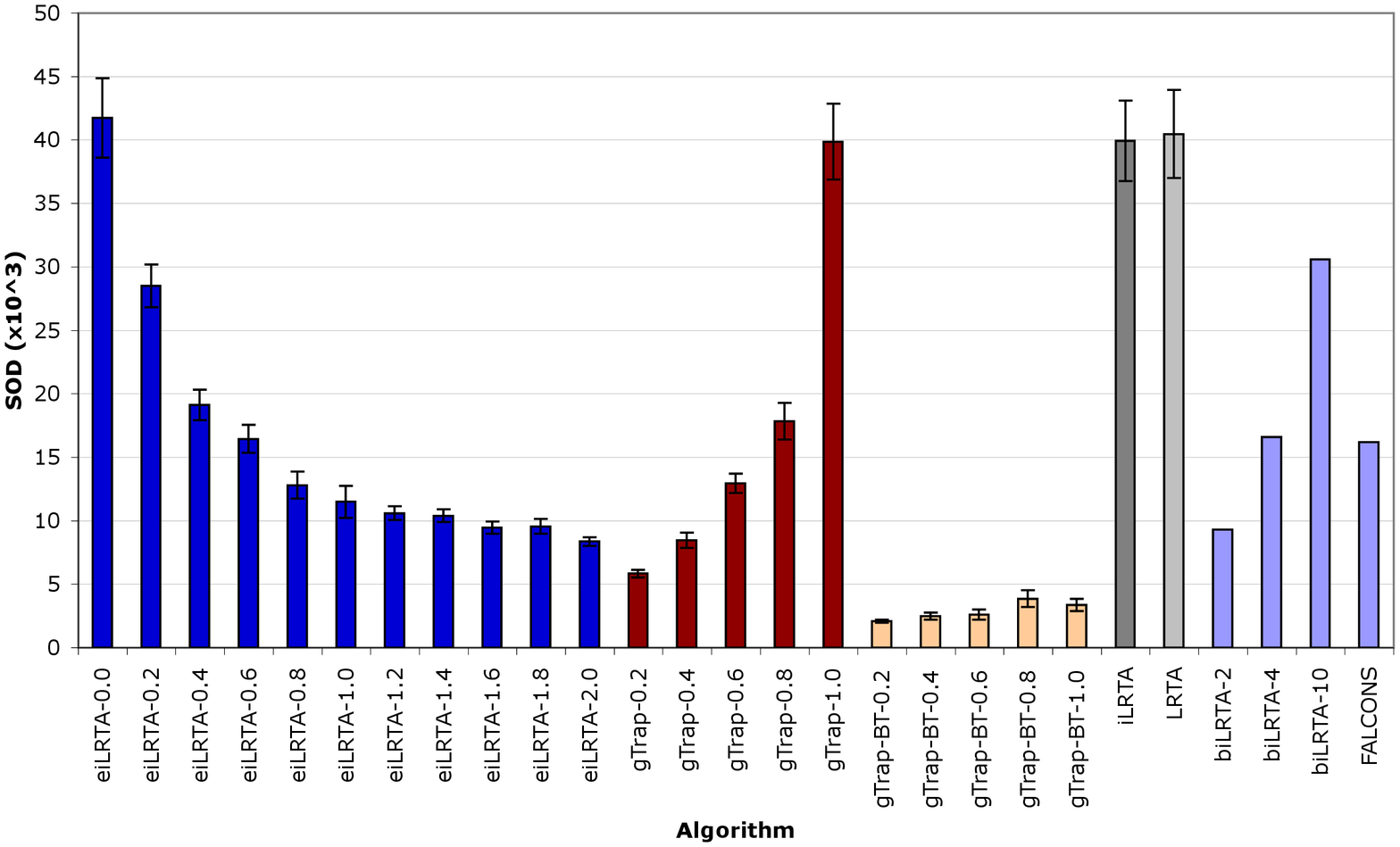}
\includegraphics[width=12.5cm]{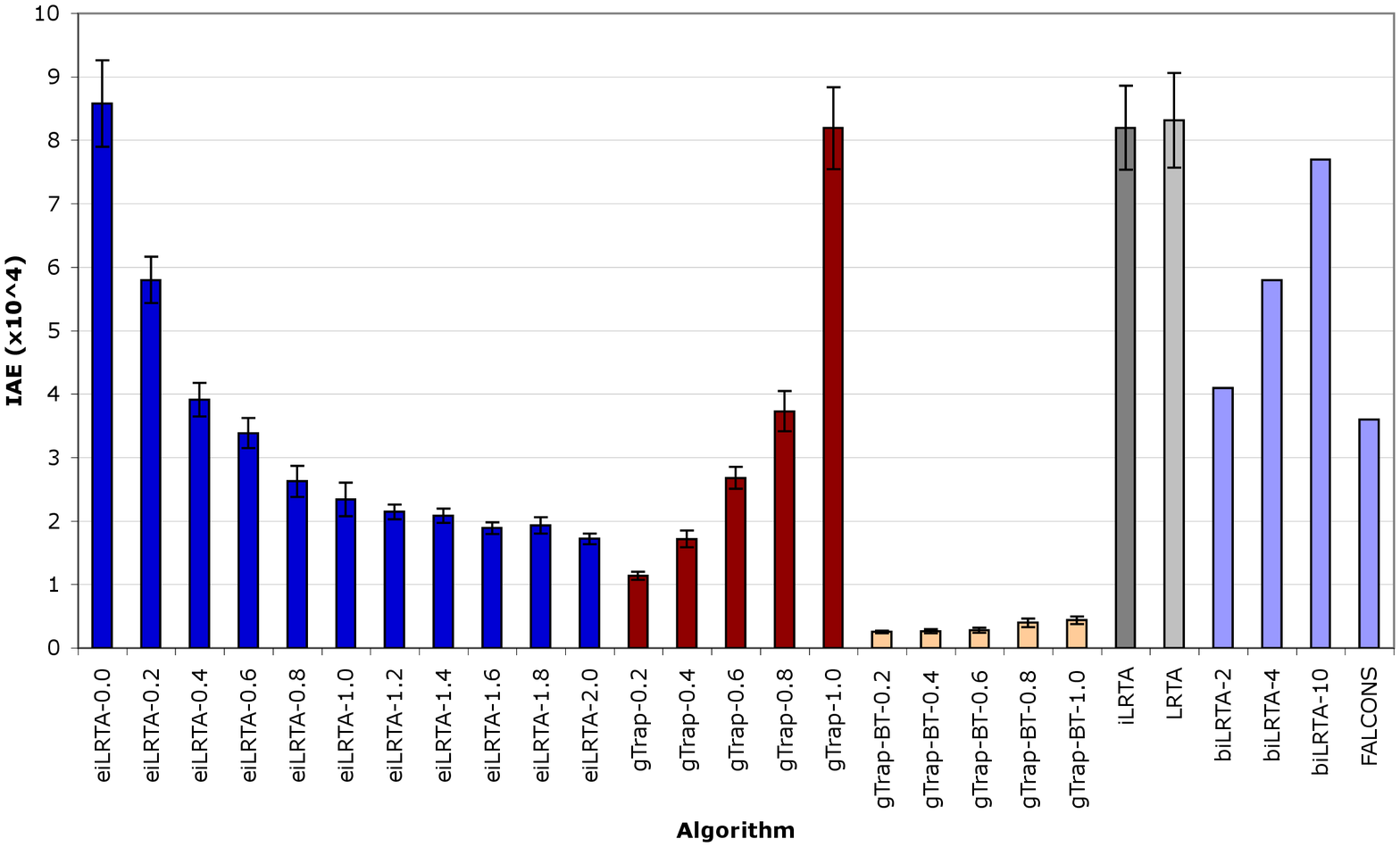}
\end{center}
\end{figure}

\medskip\nin{\bf Convergence Stability.} Even though $\gamma$-Trap was not designed to minimize the oscillations in the solution cost during the convergence process, we evaluated it against the standard algorithms along that dimension. Paralleling the study of~\cite{ShimboIshida03}, we adopted IAE, ISE, ITAE, ITSE, and SOD stability indices. IAE and SOD reported in this paper are defined as: $\text{IAE}  =   \avg_s \sum_{i=1}^{\infty} |\cost(s,i) - h^*(s)|$, $\text{SOD} =   \avg_s \sum_{i=1}^{\infty} \max\left(0,\cost(s,i+1)-\cost(s,i)\right)$ where $\cost(s,i)$ is the state $s$'s solution cost on $i$th trial while $h^*(s)$ is the optimal cost.
The 8-puzzle results for IAE and SOD with the lookahead of one are plotted in Figure~\ref{fig:8stab-1}. Similar improvements were observed for deeper lookaheads. The data for bounded LRTA* and FALCONs are taken from~\cite{ShimboIshida03} with the proper scaling factors of: $10^4$ for IAE, $10^7$ for ISE, $10^6$ for ITAE, $10^9$ for ITSE, and $10^3$ for SOD~\cite{ShimboPC}. We observe that $\gamma$-Trap with backtracking learns in a significantly more stable fashion than all other algorithms including bounded LRTA* specifically designed for stable convergence. In particular, for $\gamma=0.2$ it breaks the past records by nearly 5 folds (in SOD) and over 14 folds (in IAE). $\gamma$-Trap without backtracking appears comparable to weighted LRTA*.

\begin{figure}
\caption{First-trial solution quality with the lookahead of one. The top graph is for 10 folds of 100 random 8-puzzles each. The bottom graph is for one fold of Korf's one hundred 15-puzzles.}\label{fig:815f-1}
\begin{center}
\includegraphics[width=12.5cm]{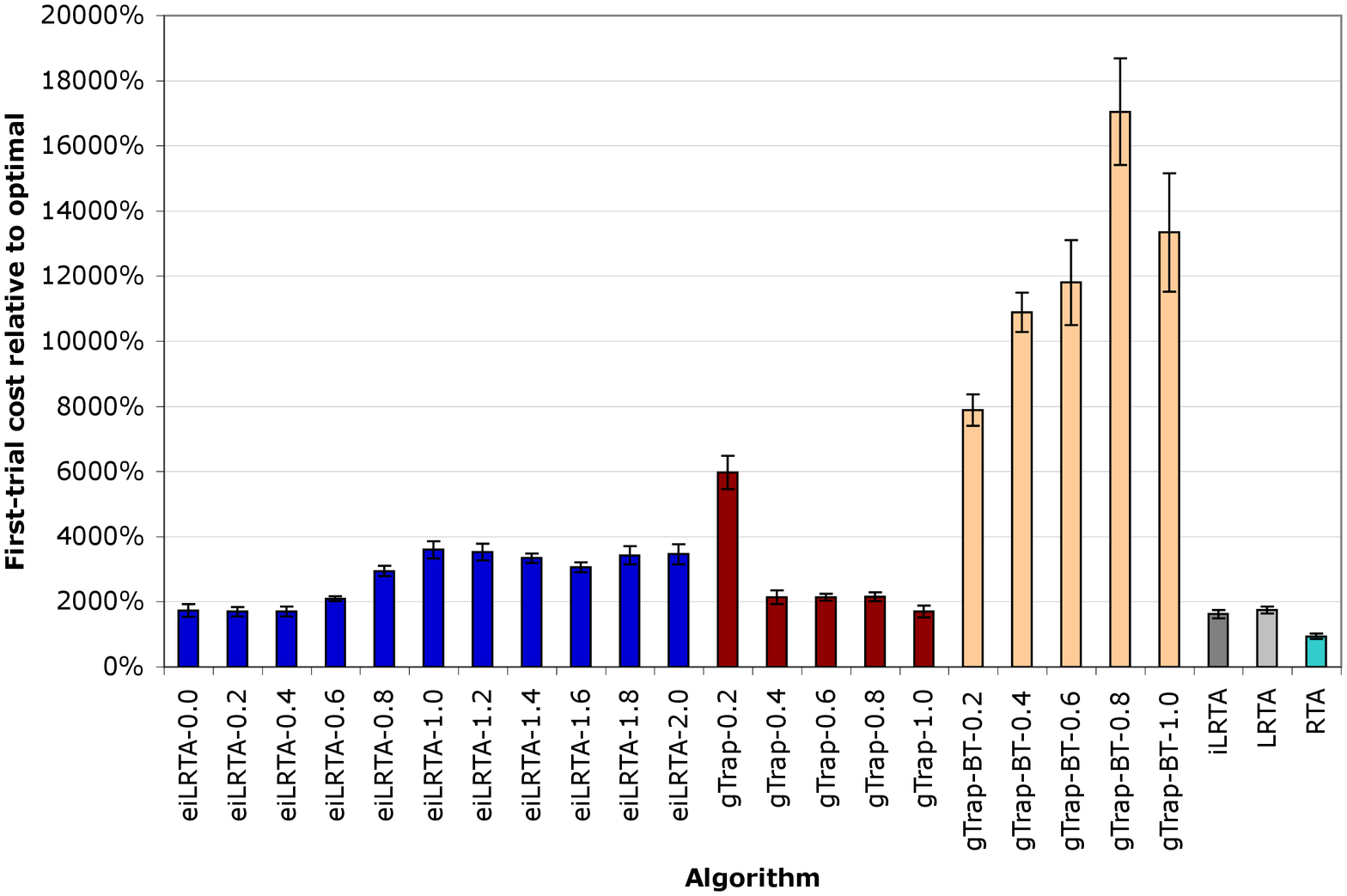}
\includegraphics[width=12.5cm]{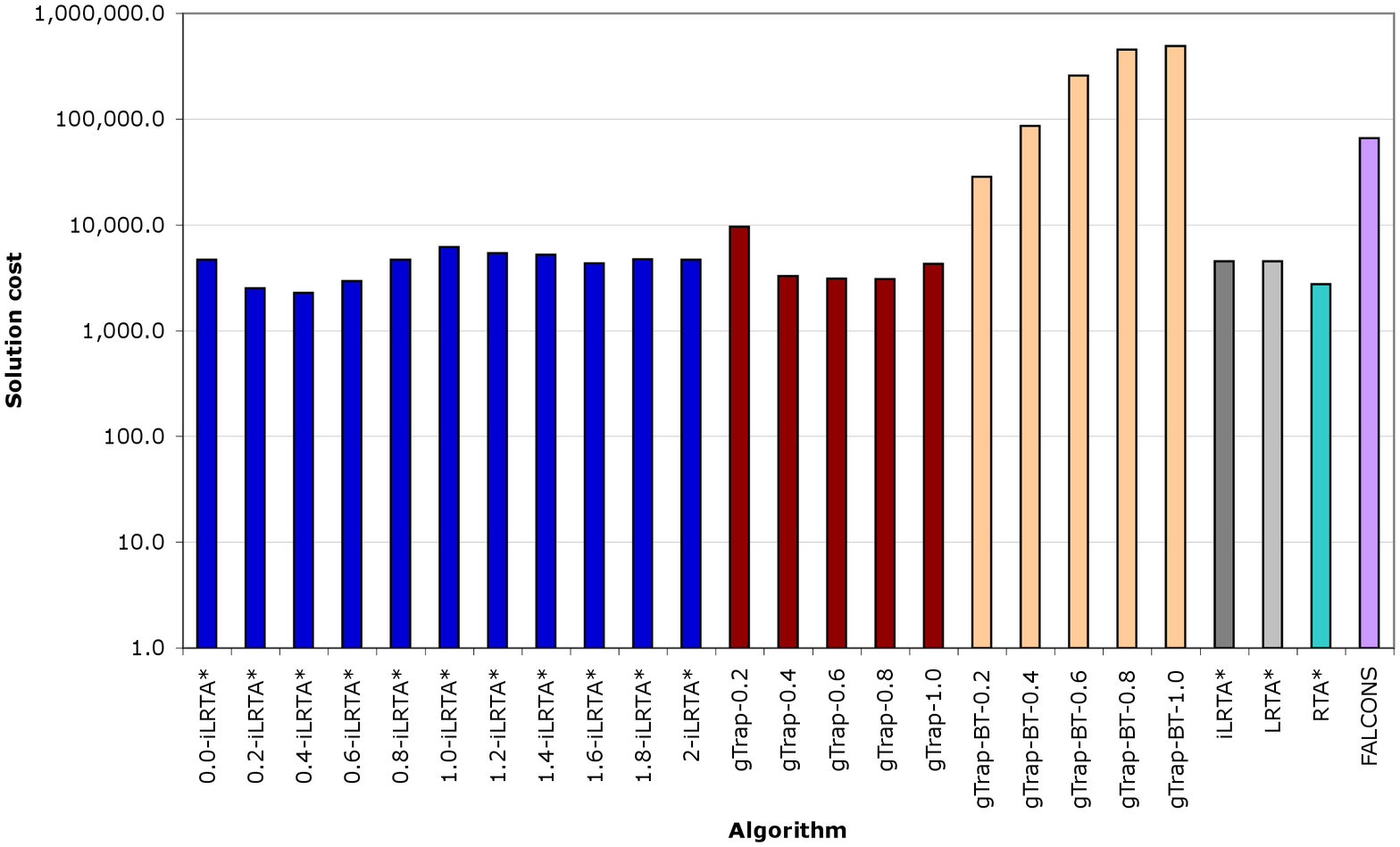}
\end{center}
\end{figure}

\medskip\nin{\bf First-trial Performance.} We evaluated the first-trial performance of the LRTS algorithms on random sets of 8-puzzles and Korf's set of 15-puzzles. Namely, within a single fold, each algorithm was presented with 100 puzzles each of which it had to solve within 500 thousand moves. The experiments were repeated over 10 independent folds each with the lookahead of 1, 2, 5, 10, and 15. The results are plotted in Figure~\ref{fig:815f-1}. We included the data reported by Shimbo and Ishida in~\cite{ShimboIshida03,ShimboPC} in the bottom graph.
The data suggest that the superior convergence speed, memory requirements, and learning stability of $\gamma$-Trap come at the cost of a more expensive first-trial solution. This behavior is similar to that exhibited by FALCONS. Additionally, we note that the backtracking part of $\gamma$-Trap appears to be the primary factor of its superior learning and inferior first-trial performance as the no-backtracking version (gTrap) is similar to the weighted LRTA* in both respects.

\vspace{-0.4cm}
\section{Future Work \& Conclusions}
\vspace{-0.4cm}
 
Learning real-time search is an appealing model for the study of rational autonomous decision-making agents. The real-time nature allows to test various strategies of interleaving planning and acting while the learning feature enables the agent to cope with unknown, uncertain, and non-stationary environments. 
Complete search methods, such as IDA*, face the intractability problem apparent even in toy domains (e.g., the scalable sliding tile puzzle). Real-time search agents trade intractable optimal solutions for effectively computable satisfactory solutions. In this paper we discuss important attributes of a learning real-time search agent and the associated trade-offs. We then examine the existing methods including LRTA* and its extensions as well as FALCONS in the light of the desired properties. A novel algorithm, called $\gamma$-Trap, is then founded on a tighter integration of adaptive lookahead-based planning and learning modules. Under the standard assumptions, it is proved to be complete and enjoys convergence to satisfactory solutions. We then evaluate it empirically against the existing methods and show a significant improvement in the convergence speed and stability as well as much reduced memory requirements. 
Future research directions include: deriving tighter upper bounds on convergence speed, stability, and final solution cost, controlling the amount of exploration on the first trial, automatic selection of the $\gamma$ parameter, and applying $\gamma$-Trap in non-deterministic, changing, and unknown environments.

\vspace{-0.4cm}
\section*{Acknowledgements} 
\vspace{-0.2cm}
Contributions of Valeriy Bulitko, David Furcy, Masashi Shimbo, Sven Koenig, Rob Holte, Rich Korf, and Stuart Russell are much appreciated. We are grateful for the support from NSERC, the University of Alberta, and the Alberta Ingenuity Centre for Machine Learning.

\vspace{-0.4cm}{\small \bibliographystyle{splncs}
\bibliography{vb}}

\end{document}